\documentclass[conference]{IEEEtran}
\UseRawInputEncoding

\IEEEoverridecommandlockouts
\usepackage{cite}

\usepackage{longtable,booktabs}
\usepackage{diagbox}
\usepackage{multirow}
\usepackage{threeparttable}
\usepackage{amsthm,amsmath,amssymb,amsfonts}
\usepackage{bm}
\usepackage{algorithmic}
\usepackage{graphicx}
\usepackage{textcomp}
\usepackage{xcolor}
\def\BibTeX{{\rm B\kern-.05em{\sc i\kern-.025em b}\kern-.08em
    T\kern-.1667em\lower.7ex\hbox{E}\kern-.125emX}}
\begin{document}

\title{Probabilistic Forecast Reconciliation with Kullback-Leibler Divergence Regularization}

\author{\IEEEauthorblockN{Guanyu Zhang}
\IEEEauthorblockA{\textit{AI Lab} \\
\textit{Lenovo Research}\\
Beijing, China \\
zhanggy17@lenovo.com}
\and
\IEEEauthorblockN{Feng Li\IEEEauthorrefmark{2}}
\IEEEauthorblockA{\textit{School of Statistics and Mathematics} \\
\textit{Central University of Finance and Economics}\\
Beijing, China (Corresponding author)\\
feng.li@cufe.edu.cn}
\and
\IEEEauthorblockN{Yanfei Kang}
\IEEEauthorblockA{\textit{School of Economics and Management} \\
\textit{Beihang University}\\
Beijing, China \\
yanfeikang@buaa.edu.cn}
\thanks{Presented in The 2023 IEEE International Conference on Data Mining (ICDM): Artificial Intelligence for Time Series Analysis Workhsop (AI4TS).}
}
\maketitle

\begin{abstract}
As the popularity of hierarchical point forecast reconciliation methods increases, there is a growing interest in probabilistic forecast reconciliation. Many studies have utilized machine learning or deep learning techniques to implement probabilistic forecasting reconciliation and have made notable progress. However, these methods treat the reconciliation step as a fixed and hard post-processing step, leading to a trade-off between accuracy and coherency. In this paper, we propose a new approach for probabilistic forecast reconciliation. Unlike existing approaches, our proposed approach fuses the prediction step and reconciliation step into a deep learning framework, making the reconciliation step more flexible and soft by introducing the Kullback-Leibler divergence regularization term into the loss function. The approach is evaluated using three hierarchical time series datasets, which shows the advantages of our approach over other probabilistic forecast reconciliation methods.

\end{abstract}


\section{Introduction}
Multivariate time series in many practical applications are often arranged in hierarchies, i.e., time
series at upper levels of the hierarchy are aggregates of those at lower levels, which can be represented by a tree diagram. When tackling the hierarchical time series forecasting problem, for interpretability and better helping downstream tasks to make decisions, we require that the forecasts meet the coherency, i.e. forecasts of aggregated time series are the sum of forecasts of the corresponding disaggregated time series. To achieve coherency, a rich literature has emerged on forecast reconciliation, such as bottom-up (BU), top-down (TD), and MinT \cite{HYNDMAN201616}.

Currently, most reconciliation methods are based on point forecasts, and there is little research on probabilistic forecast reconciliation. However, probabilistic forecast is very important in practical applications and is helpful for risk management and decision-making. Especially in the power, wind, and other new energy industries, data is very unstable, and the probabilistic forecast is more practical than point forecast, which helps staff allocate energy rationally and maximize the efficiency of energy use \cite{ZHANG2014255}. Probabilistic forecast reconciliation is very challenging because it does not just reconcile the mean prediction values but reconciles multiple related predictive distributions. Most of the probabilistic forecast reconciliation methods proposed in existing research are realized by sampling, reordering, specifying hierarchical dependencies, and choosing reconciliation weights by optimizing a scoring rule \cite{JEON2019364}. These approaches have two drawbacks: First, the whole process contains too many artificial assumptions and is not totally data-driven. Second, the prediction and reconciliation steps are regarded as two separate independent processes, which is not conducive to parameter tuning and model updating under data mode conversion, and the reconciliation step cannot be adjusted accordingly for different prediction methods. Although Rangapuram \emph{et al.} proposes an end-to-end framework to solve this problem, the reconciliation step in this framework is actually a hard mapping step and not flexible and soft enough \cite{Rangapuram2021EndtoEndLO}.

Recently, deep learning has shown great potential in the field of time series forecasting. Deep neural networks can learn complex data representations, which reduces the need for manual feature engineering and model design. Outputs and loss functions are flexible, allowing for the customization of different aims, such as point forecasting, probabilistic forecasting, and other special requirements. Some novel deep learning architectures developed for time series forecasting include DeepAR \cite{SALINAS20201181}, N-BEATS \cite{DBLP:journals/corr/abs-1905-10437}, and Temporal Fusion Transformer \cite{LIM20211748}. However, the application of deep learning in the field of probabilistic forecast reconciliation is still lacking.

Inspired by the end-to-end method proposed by Rangapuram \emph{et al.} \cite{Rangapuram2021EndtoEndLO}, we propose a new probabilistic forecast reconciliation method based on deep learning to solve the above problems. First, we use a deep learning model to predict multivariate time series and introduce the Kullback–Leibler divergence (KLD) regularization term into the loss function, which measures the ``distance" between multiple distributions, i.e. the distance between the parent node distribution and the sum of the child node distributions. In this way, the prediction and reconciliation steps are fused into a deep learning framework, instead of treating reconciliation as a post-processing step, the KLD regularization term will help the model achieve approximate coherency of the predictive distribution at training time. We set a coefficient for the KLD regularization term to control the degree of reconciliation, which can be tuned as a hyperparameter based on different objectives. For example, we can select the optimal coefficient based on the accuracy of a validation set to avoid sacrificing accuracy for the sake of coherency. Second, the multivariate predictive distribution given by the trained model does not meet the complete coherency, so the idea of bottom-up can be used to transform the incompletely coherent multivariate distribution into a strictly coherent multivariate distribution.

Compared with the existing probabilistic forecast reconciliation methods, our proposed method has the following advantages: First, the prediction step and reconciliation step are integrated into a deep learning framework, which is conducive to direct optimization of the target, avoiding error accumulation and facilitating model updating. Second, MinT and other traditional reconciliation methods have limited consideration of the correlation between multiple time series. However, the deep learning model is more able to capture relevant information. Third, the coefficient in the KLD regularization term is adjustable, which is conducive to adjusting the degree of reconciliation. Sometimes, there is a trade-off between coherency and accuracy, and we can set a reasonable degree of reconciliation to meet different research goals.

The rest of the paper is organized as follows. In Section \ref{sec:Related work}, we introduce the notation and review the existing common forecast reconciliation methods. We then demonstrate our new probabilistic forecast reconciliation method in Section \ref{sec:pfr_dl}. In Section \ref{sec:Experiments}, we conduct experiments on three data sets to compare our approach with existing common probabilistic forecast reconciliation methods. We conclude the paper and discuss future work in Section \ref{sec:conclusions}. The code for reproducing the results is
available at https://github.com/guanyu0316/Probabilistic-Forecast-Reconciliation-with-DL.

\section{Preliminaries}
\label{sec:Related work}
\subsection{Key Terms and Definitions}
\begin{figure}
  % \begin{center}
    \begin{tabular}{c}
      \includegraphics[width=8cm]{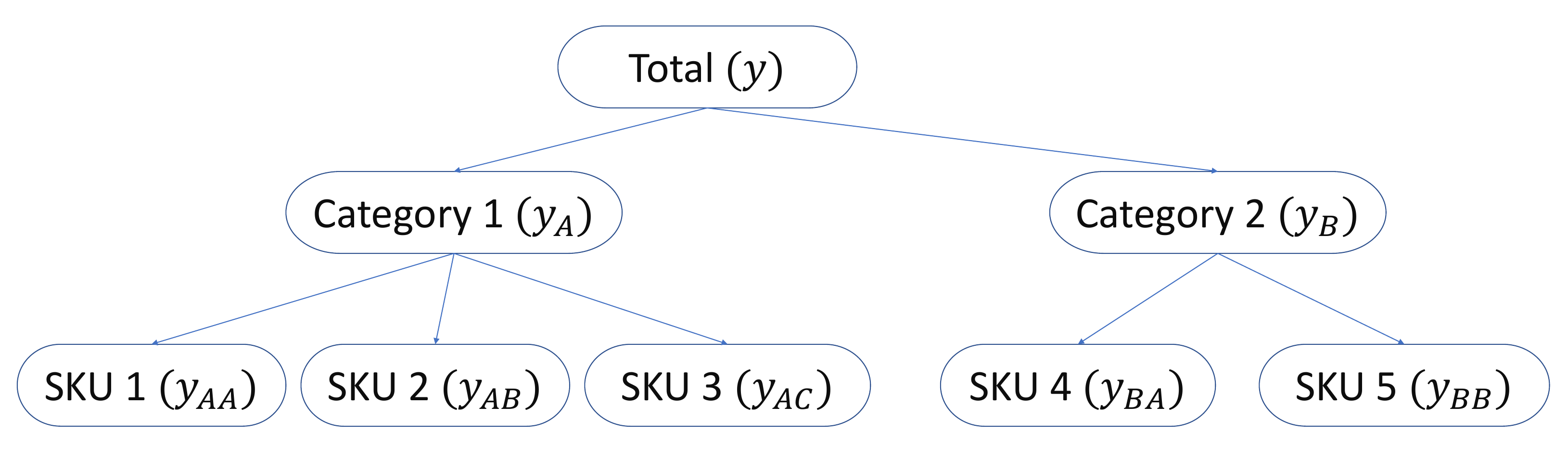}
    \end{tabular}
  % \end{center}
  \caption
  { \label{hier-example}
    A hierarchy example. }
\end{figure}

Let $\boldsymbol{y}_{\boldsymbol{t}}$ denote an n-vector containing observations at time $t$ of all nodes in a hierarchical structure. Let $\boldsymbol{b}_{\boldsymbol{t}}$ be an m-vector with the observations
at time $t$ of bottom nodes. In the field of hierarchical forecasting, an important matrix is the summing matrix $\boldsymbol{S}$, which is able to map the bottom time series to $\boldsymbol{y}_{\boldsymbol{t}}$, i.e. $\boldsymbol{y}_{\boldsymbol{t}}=\boldsymbol{S}\boldsymbol{b}_{\boldsymbol{t}}$. Once the hierarchical structure is determined, $\bm{S}$ is determined. Taking the hierarchical structure shown in Figure \ref{hier-example} as an example, a company can organize its five SKUs into two categories, then

\setlength{\arraycolsep}{0.8pt}
$$\left[\begin{array}{c}
y_h \\
y_{A, h} \\
y_{B, h} \\
y_{A A, h} \\
y_{A B, h} \\
y_{A C, h} \\
y_{B A, h} \\
y_{B B, h}
\end{array}\right]=\left[\begin{array}{lllll}
1 & 1 & 1 & 1 & 1 \\
1 & 1 & 1 & 0 & 0 \\
0 & 0 & 0 & 1 & 1 \\
1 & 0 & 0 & 0 & 0 \\
0 & 1 & 0 & 0 & 0 \\
0 & 0 & 1 & 0 & 0 \\
0 & 0 & 0 & 1 & 0 \\
0 & 0 & 0 & 0 & 1
\end{array}\right]\left[\begin{array}{l}
y_{A A, h} \\
y_{A B, h} \\
y_{A C, h} \\
y_{B A, h} \\
y_{B B, h}
\end{array}\right].$$

The goal of point forecast is to estimate the $h$ period-ahead conditional expectation
$${E}\left[\boldsymbol{y}_{T+h} \mid \boldsymbol{y}_1, \ldots, \boldsymbol{y}_T\right].$$
We can compute forecasts for all series at all levels, which we call base forecasts $\hat{\boldsymbol{y}}_h$. However, these forecasts may not meet coherency, i.e. $\hat{\boldsymbol{y}}_h=\boldsymbol{S}\hat{\boldsymbol{b}}_h$, so we need to transform the base forecasts to meet the coherency, and this process is called reconciliation. $\tilde{\boldsymbol{y}}_h$ represents forecast after being reconciled, which must meet coherency.

While the goal of probabilistic forecast is to estimate, for each node in the hierarchy, the $h$ period-ahead conditional cumulative distribution function
$$F_{i, T+h}\left(y \mid \boldsymbol{y}_1, \ldots, \boldsymbol{y}_T\right)={P}\left(y_{i, T+h} \leq y \mid \boldsymbol{y}_1, \ldots, \boldsymbol{y}_T\right).$$ In the field of probabilistic forecast, coherency is defined as follows: Let $X_i \sim \hat{F}_i \ for \ i=1,...,n$ and let $i_1, \ldots, i_{n_k}$ denote the $n_{k}$ children of series $i$. The forecasts $\hat{F}_i$ are probabilistic coherent if $X_i \stackrel{d}{=} X_{i_1}+\cdots+X_{i_{n_k}} \ for \ i=1,...,r$, where $\stackrel{d}{=}$ denoted equality in distribution \cite{Souhaib}. In other words, the forecast is coherent if the predictive distribution of each aggregate series is equal to the distribution of the sum of the children series. The process of generating coherent probabilistic forecasts using a specific method is referred to as probabilistic forecast reconciliation, which is the focus of this study.

\subsection{Benchmark Models}
In this subsection, we will introduce the benchmark models used for comparison. For illustrative purposes, we take the hierarchical structure shown in Figure \ref{hier-example} as an example. In order to clearly introduce the existing probabilistic forecast reconciliation methods, let's first introduce point forecast reconciliation methods. Common point forecast reconciliation methods include bottom-up, top-down, and MinT. The essence of these reconciliation methods is the linear transformation
$$
\widetilde{\bm{y}}_h=\bm{S} \bm{P} \widehat{\bm{y}}_h.
$$
The $\bm{P}$ matrix can map the base forecast $\widehat{\bm{y}}_h$ to the forecast on the bottom level, then $\bm{S}$ can restore this bottom level forecast to the forecast vector of all nodes. Once the hierarchical structure is determined, $\bm{S}$ is determined, while $P$ depends on different reconciliation methods. For example, in the bottom-up approach
$$
\setlength{\arraycolsep}{0.8pt}
\boldsymbol{P}=\left[\begin{array}{llllllll}
0 & 0 & 0 & 1 & 0 & 0 & 0 & 0 \\
0 & 0 & 0 & 0 & 1 & 0 & 0 & 0 \\
0 & 0 & 0 & 0 & 0 & 1 & 0 & 0 \\
0 & 0 & 0 & 0 & 0 & 0 & 1 & 0 \\
0 & 0 & 0 & 0 & 0 & 0 & 0 & 1
\end{array}\right],
$$
i.e., the reconciled forecast is obtained by aggregating the bottom-level forecast. In the top-down method
$$
\setlength{\arraycolsep}{0.8pt}
\boldsymbol{P}=\left[\begin{array}{llllllll}
p_1 & 0 & 0 & 0 & 0 & 0 & 0 & 0 \\
p_2 & 0 & 0 & 0 & 0 & 0 & 0 & 0 \\
p_3 & 0 & 0 & 0 & 0 & 0 & 0 & 0 \\
p_4 & 0 & 0 & 0 & 0 & 0 & 0 & 0 \\
p_5 & 0 & 0 & 0 & 0 & 0 & 0 & 0
\end{array}\right],
$$
i.e. the forecast of the top node is disassembled to the bottom level according to a certain proportion vector, and the bottom level forecast will then be aggregated into the forecast of all nodes using $\bm{S}$. Wickramasuriya \emph{et al.} proved that when $\bm{P}$ takes the next formula, the trace of the variance-covariance matrix $\boldsymbol{V}_h$ of the reconciled prediction error is the smallest \cite{Wickramasuriya2019Optimal}. Therefore, the optimal reconciliation method is also called MinT (Minimize Trace).
$$
\boldsymbol{P}=\left(\boldsymbol{S}^{\prime} \boldsymbol{W}_h^{-1} \boldsymbol{S}\right)^{-1} \boldsymbol{S}^{\prime} \boldsymbol{W}_h^{-1}, \\
$$
$$
\boldsymbol{V}_h=\operatorname{Var}\left[\boldsymbol{y}_{T+h}-\widetilde{\boldsymbol{y}}_h\right]=\boldsymbol{S} \boldsymbol{P} \boldsymbol{W}_h \boldsymbol{P}^{\prime} \boldsymbol{S}^{\prime},
$$
where $\boldsymbol{W}_h=\operatorname{Var}\left[\left(\boldsymbol{y}_{T+h}-\widehat{\boldsymbol{y}}_h\right)\right]$ is the variance-covariance matrix of base forecast error, which is unknown and difficult to be estimated accurately. In practice, it is common to make simplifying assumptions about $\boldsymbol{W}_h$. A commonly used assumption is $\boldsymbol{W}_h=k_h \boldsymbol{\Lambda}, \boldsymbol{\Lambda}=\operatorname{diag}(\mathbf{S 1})$, i.e. the base forecast error variance of a node is proportional to the number of child nodes it contains, which is called STRUCT assumption.

Compared with the research on point forecast reconciliation, few studies on probabilistic forecast reconciliation exist. Jeon \emph{et al.} proposed a relatively complete probabilistic forecast reconciliation approach \cite{JEON2019364}. Specifying the hierarchical sample arrangement form, the point forecast reconciliation methods can be applied to probabilistic forecast reconciliation. Suppose we can obtain a sample from the predictive joint distribution:
$$
    \widehat{\boldsymbol{Y}}=\left(\widehat{\boldsymbol{y}}_1^{t+h \mid t}, \ldots, \widehat{\boldsymbol{y}}_N^{t+h \mid t}\right).
$$
Every column of $\widehat{\boldsymbol{Y}}$ is a sample vector from the predictive joint distribution. We can apply the formula of point forecast reconciliation to probabilistic forecast as follows. Point forecast reconciliation is to reconcile a forecast vector, while probabilistic forecast reconciliation is to reconcile a matrix:
$$
    \bm{\widetilde{Y}=S P \widehat{Y}s}.
$$

Jeon \emph{et al.} proposed three ways to construct joint distribution samples, which are abbreviated as stack, rank and random \cite{JEON2019364}. The stack method refers to that we sample from the predictive distribution of all nodes separately, and then the samples are concatenated directly:
$$
\widehat{\boldsymbol{Y}}^S=\left[\begin{array}{c}
\widehat{\mathbf{Z}}_1 \\
\widehat{\mathbf{Z}}_2 \\
\vdots \\
\widehat{\boldsymbol{Z}}_L
\end{array}\right],
$$
where $\widehat{\boldsymbol{Z}}_l$ is vector of $N$ samples from predictive distribution of node $l$. The essence of this approach is to assume that the time series of all nodes are independent. The rank method means that the values of column $i$ of the constructed joint distribution sample $\widehat{\boldsymbol{Y}}^S$ are all the $(i/N)$ quantiles of the predictive distributions of all nodes. The rank method assumes that the series of all nodes or all levels are highly correlated. The random method refers to the random rearrangement of the samples drawn from each node and then concatenating them together like the stack method. The implicit assumption of the random method is that the series of all nodes are weakly correlated.

After obtaining the joint distribution samples through different permutation methods, we can use MinT or BU methods to reconcile the predictive joint distribution samples just like point forecast reconciliation, which can be understood as treating each column of $\widehat{\boldsymbol{Y}}$ as a point forecast sequence and reconciliating them separately.

Currently, both the point forecast reconciliation method and the probabilistic forecast reconciliation method can be regarded as a two-step process, i.e., the first step is to obtain the base forecast, and the second is to reconcile the base forecast. This idea is not conducive to tuning parameters and model adjustment under data mode transformation. In addition, the above probabilistic forecasting reconciliation method implies too many artificial assumptions of hierarchical dependence and is not flexible enough.

\section{Probabilistic Forecast Reconciliation with Kullback-Leibler Divergence Regularization}
\label{sec:pfr_dl}
In this section, we present a probabilistic forecast reconciliation approach based on deep learning with Kullback-Leibler divergence regularization, with the objective of overcoming some of the aforementioned limitations of the existing probabilistic forecast reconciliation method. Our approach can realize end-to-end training and ensure the coherency of probabilistic forecast results. In our approach, the deep learning method is not specified, all global probabilistic forecasting models for multivariate time series based on deep learning can be used in our approach.
\subsection{Proposed approach}
Taking the hierarchical structure in Figure \ref{hier-example} as an example, there are eight time series in total, and the value of the time series $i$ at time $t$ is defined as $y_{i,t}$, $[1,t_0-1]$ is the history period or conditioning range, $[t_0,T]$ is the prediction range. Suppose we use a deep learning model for producing probabilistic forecasts, which can model multiple time series simultaneously. In this deep learning model, the input to the network is history information and available covariates, and the output is the parameter of the predictive probabilistic distribution. Given a data set of time series and associated covariates, the parameters of this deep learning model can be learned by minimizing the loss:
$$
L=\sum_{i=1}^N \sum_{t=t_0}^T l\left(y_{i, t},\theta_{i, t}\right).
$$
Where $\theta_{i, t}$ is the parameter of the predictive probabilistic distribution, and N is 8 in the example of Figure \ref{hier-example}. $l$ can be the negative log-likelihood function. In order to meet the coherency of multiple series probabilistic forecasts, we add a KLD regularization term to the loss function:
$$
L=\sum_{i=1}^N \sum_{t=t_0}^T l\left(y_{i, t},\theta_{i, t}\right) + KL_{reg}.
$$
Taking the hierarchical structure in Figure \ref{hier-example} as an example, $KL_{reg}$ can be represented by the following equation:
$$
\begin{gathered}
KL_{r e g}=\lambda_1 D_{K L}\left(F_{Y_{. t}} \| F_{\boldsymbol{Y}_{A . t}+\boldsymbol{Y}_{B . t}}\right)\\
+\lambda_2 D_{K L}\left(F_{\boldsymbol{Y}_{A . t}} \| F_{\boldsymbol{Y}_{A A . t}+\boldsymbol{Y}_{A B . t}+\boldsymbol{Y}_{A C . t}}\right) \\
+\lambda_3 D_{K L}\left(F_{\boldsymbol{Y}_{B . t}} \| F_{\boldsymbol{Y}_{B A . t}+\boldsymbol{Y}_{B B . t}}\right),
\end{gathered}
$$
where $\boldsymbol{Y}_{i, t} \sim F_{\boldsymbol{Y}_{i, t}}$. Multiple KLD terms are added, and each term is the KLD between the distribution of a parent node and the distribution of the sum of its child nodes. In mathematical statistics, the KLD denoted $D_{KL}(P \| Q)$, measuring how one probabilistic distribution $P$ differs from a second, reference probabilistic distribution $Q$. The specific definition of KLD is: for discrete probabilistic distributions $P$ and $Q$ defined on the same sample space, the KLD from $Q$ to $P$ is defined to be
$$
D_{KL}(P \| Q)=\sum_{x \in \mathcal{X}} P(x) \log \left(\frac{P(x)}{Q(x)}\right).
$$
For distributions $P$ and $Q$ of a continuous random variable, KLD is defined to be the integral:
$$
D_{KL}(P \| Q)=\int_{-\infty}^{\infty} p(x) \log \left(\frac{p(x)}{q(x)}\right) d x.
$$
So KLD is not symmetric in the two distributions. When calculating KLD, we swap positions between the two distributions and calculate the mean of the two KLDs. Taking the first item in $KL_{reg}$ as an example, let's amend it to
$$
\begin{gathered}
D_{symmetric}=\\
[D_{K L}\left(F_{Y, t} \| F_{Y_{A, t}+Y_{B, t}}\right)+D_{K L}\left(F_{Y_{A, t}+Y_{B, t}} \| F_{Y_{, t}}\right)] / 2.
\end{gathered}
$$

The KLD regularization term will help the model achieve approximate coherency of the predictive probabilistic distribution at training time. The point is how to guarantee the differentiability of the regularization term in the neural network. KLD is usually calculated using samples from the two distributions. In order to ensure the differentiability of the sampling step, we use the reparameterization technique to make the sampling step independent of the model, so that the sample from the two distributions is only related to the distribution parameters, i.e. only related to parameters in the whole neural network, which ensures the differentiability of this step. For example, suppose the predictive distribution is Gaussian distribution. Before training, we draw samples from the standard multivariate normal distribution, i.e. $\bm{z \sim N(0, I)}$. When we want to sample from the predictive distribution, we do the reparameterization:
$$
\bm{y_t=\mu_t+\Sigma_t^{1 / 2} z},
$$
then the sampling step is outside of the network, which ensures the differentiability. When calculating the distribution of the sum of child nodes, we assume that the time series of each child node is independent of each other, but this does not mean that our method does not consider the correlation information between each time series, because the deep learning method itself has taken into account the correlation, which is the advantage of deep learning.

The $\lambda$ coefficient in the KLD regularization term is used to control the degree of reconciliation. In practice, coherency usually improves accuracy \cite{Wickramasuriya2019Optimal}, and subsequent experiments in this paper confirm this. However, for deep learning models, adding regularization terms to the loss function may affect the accuracy, and there is a trade-off between accuracy and coherency, so we set $\lambda$ as a hyperparameter. $\lambda$ close to zero does not enforce coherency, while larger $\lambda$ make predictions more coherent. Depending on the practical application, we can choose a soft penalty version or a hard constraint version. In our experiment, we set the same $\lambda$ for all items in $KL_{reg}$. However, different $\lambda$ can be set if having sufficient GPU computing power.

In the actual training process, it is necessary to specify the batch size as an integer multiple of the total number of nodes in the hierarchical structure, i.e. each training batch must exactly contain the time series of all nodes so that the KLD regularization term can be calculated. We need to be careful to generate multiple training instances and mark the node corresponding to each instance, which facilitates the construction of batches and the calculation of loss. The forecast given by the above model after training is not entirely coherent, because the penalty term only helps the model give an approximately coherent forecast. One more step of bottom-up aggregation is required to give coherent forecasts.

The following subsection presents one deep learning model, DeepAR, in this study for reconciliation. The basic idea of DeepAR, training mode, and details of implementation will be given.

\subsection{DeepAR}
DeepAR is a time series probabilistic forecast method based on deep learning proposed by Flunkert \emph{et al.} \cite{SALINAS20201181}. It has been applied to various forecasting problems with excellent results. DeepAR is a nonlinear form of the classical autoregressive model, which can be applied to single time series or extended to multiple time series. DeepAR is based on Recurrent Neural Network (RNN). The objective of DeepAR is to obtain the joint conditional probabilistic distribution $P\left(\boldsymbol{y}_{i, t_0: T} \mid \boldsymbol{y}_{i, 1: t_0-1}, \boldsymbol{x}_{i, 1: T}\right)$, i.e. according to the existing data $\boldsymbol{y}_{i, 1: t_0-1}$ and covariates $\boldsymbol{x}_{i, 1: T}$, modeling the future sequence $\boldsymbol{y}_{i, t_0: T}$. The above distribution can be written in the following likelihood form:
$$
\begin{gathered}
Q_{\Theta}\left(\boldsymbol{y}_{i, t_0: T} \mid \boldsymbol{y}_{i, 1: t_0-1}, \boldsymbol{x}_{i, 1: T}\right)=\prod_{t=t_0}^T Q_{\Theta}\left(y_{i, t} \mid \boldsymbol{y}_{i, 1: t_0-1}, \boldsymbol{x}_{i, 1: T}\right) \\
=\prod_{t=t_0}^T \ell\left(y_{i, t} \mid \theta\left(\mathbf{h}_{i, t}, \Theta\right)\right),
\end{gathered}
$$
where $\mathbf{h}_{i, t}=h\left(\mathbf{h}_{i, t-1}, y_{i, t-1}, \boldsymbol{x}_{i, t}, \Theta\right)$  is the output of RNN. We input the hidden state $\mathbf{h}_{i, t-1}$ at the last moment and data $y_{i, t-1}$, as well as covariate $\boldsymbol{x}_{i, t}$, then RNN will output $\mathbf{h}_{i, t}$, which is then transformed into the parameters of a given distribution by a specific transformation $\theta(\cdot)$, and then the likelihood function is obtained. In the experiment of this paper, we use Gaussian distribution to calculate the likelihood function. The mean value $\mu$ of Gaussian distribution is given by the affine function of network output, and the standard deviation $\sigma$ is obtained by applying affine transformation and softplus activation function to ensure $\sigma>0$, i.e.
$$
\mu\left(\mathbf{h}_{i, t}\right)=\mathbf{w}_\mu^T \mathbf{h}_{i, t}+b_\mu, \sigma\left(\mathbf{h}_{i, t}\right)=\log \left(1+\exp \left(\mathbf{w}_\sigma^T \mathbf{h}_{i, t}+b_\sigma\right)\right).
$$
Then the negative log-likelihood loss function is as follows:
$$
L=-\sum_{i=1}^N \sum_{t=t_0}^T \log \ell\left(y_{i, t} \mid \theta\left(\mathbf{h}_{i, t}\right)\right).
$$
We use multivariate DeepAR, so the likelihood function is multiplied by the likelihood function of all nodes. When applying DeepAR to our proposed reconciliation method, we add KLD regularization term to this loss function.
\begin{figure}
  \begin{center}
    \begin{tabular}{c}
      \includegraphics[width=8cm]{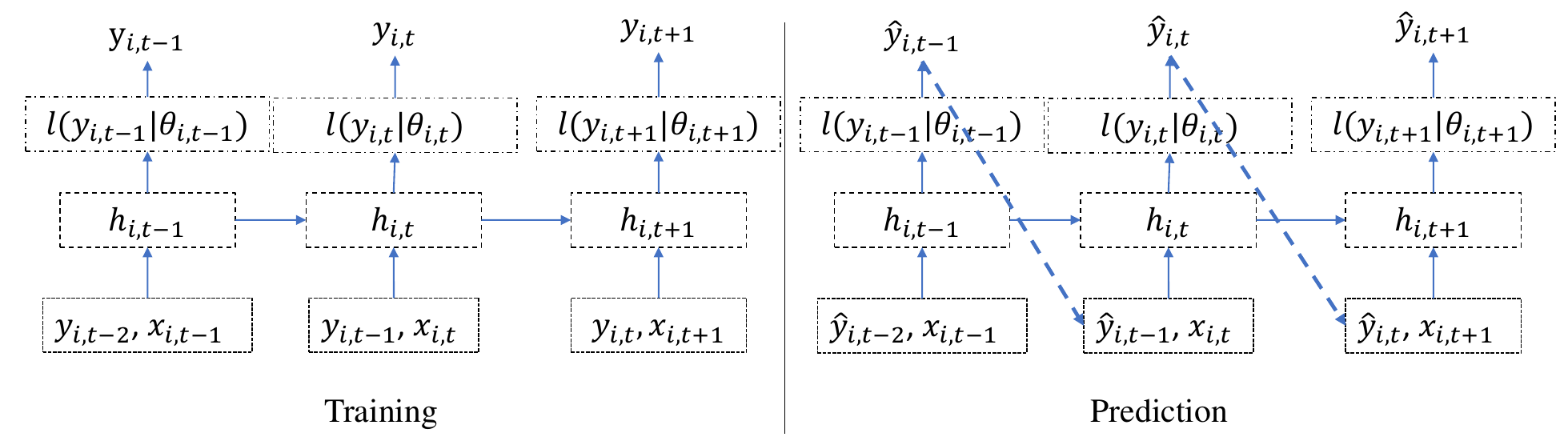}
    \end{tabular}
  \end{center}
  \caption
  { \label{deepar}
    DeepAR training and prediction process.}
\end{figure}

The training and prediction process of DeepAR can be represented by Figure \ref{deepar}. There is a difference between the training process and the prediction process, the label values are known at the time of training, so each network input is the time lag term of the real value. For the prediction process, the input of each round of the network is derived from the mean of the distribution of the previous round of network output.

In subsequent experiments, we apply DeepAR to our proposed reconciliation method for probabilistic forecasting and name it DeepAR-Hier.

\section{Experiments}
\label{sec:Experiments}
\subsection{Data}
In order to verify the effectiveness of the proposed approach, we conduct experiments on three classical data sets in the field of hierarchical forecast, i.e. the Australian infant mortality data set, the Australian tourism data set, and the Wikipedia page view data set. The background and features of these data sets are detailed below.

The Australian infant mortality data set, abbreviated as Infant, is from the Australian Social Science Data Archive and includes infant mortality rates across eight Australian states: New South Wales (NSW), Victoria (VIC), Queensland (QLD), South Australia (SA), Western Australia (WA), Northern Territory (NT), Australian Capital Territory (ACTOT), and Tasmania (TAS). This data set is grouped by gender and state. The time range of available data is from 1901 to 2003. The data set is publicly available in R's addb package \cite{addb2010}. Due to missing values, we use annual data from 1933 to 2003 in our analysis, and the forecast horizon is 4.

The Australian tourism data set, abbreviated as Tourism, contains overnight trips in Australia from the first quarter of 1998 to the fourth quarter of 2016 \cite{49018680cf534023b6ae0f4e933c475d}. The data are quarterly and grouped by three variables: tourism region, eight Australian states, and purpose of travel. For this data set, the forecast horizon is set to 4.

Wikipedia page views data set, abbreviated as Wiki, gives the daily page views of 145,000 different Wikipedia articles from 2015-07-01 to 2016-12-31, from which we sample 32 bottom time series and aggregate them to get the upper time series. Specifically, the data are grouped by three variables: agent type, access type, and country code. In the experiment, the forecast horizon is set to 15 days.

\subsection{Experimental set-up}
We choose DeepAR as the deep learning model in our proposed approach. We conducted a grid search for some important hyperparameters of DeepAR-Hier, including the hidden recurrent size between (10,30), the number of RNN layers between (2,5), dropout rate between (0.1,0.2), the number of epochs (20,60) and the coefficient of the KLD regularization term $\lambda$ between(0,1). According to the time length of different data sets, we use cross validation with a specific number of folds to search parameters.

\begin{table}[!h]
\small
  \centering
  \caption{All methods available for comparison.}
  \label{tab: comp}
    \begin{tabular}{ll}
      \toprule
      DeepAR-Hier&Arima-Random-Bu-None\\
      Pure-DeepAR&Ets-Stack-MinT-Struct \\
      Arima-Stack-MinT-Struct&Ets-Stack-Bu-None\\
      Arima-Stack-Bu-None&Ets-Rank-MinT-Struct\\
      Arima-Rank-MinT-Struct&Ets-Rank-Bu-None \\
      Arima-Rank-Bu-None&Ets-Random-MinT-Struct \\
      Arima-Random-MinT-Struct&Ets-Random-Bu-None \\
      \bottomrule
    \end{tabular}
\end{table}

We compare the proposed approach with the existing probabilistic forecast reconciliation method mentioned in section \ref{sec:Related work}. In a word, there are 14 models participating in the comparison summarised in Table \ref{tab: comp}. For example, Arima-Stack-MinT-Struct means that the base forecast is given by Arima, the joint distribution sample is constructed by stack, and the reconciliation method is MinT in which error covariance is constructed by STRUCT. The BU method does not need to assume the error covariance, which is represented by None. Pure-DeepAR refers to the original DeepAR method without changing the loss function. We use PyTorch to implement DeepAR-Hier and Pure-DeepAR, use Python's Sktime library to implement Arima and ETS, and automatically tune the model parameters of Arima and ETS.

We evaluate the forecasting performance of these methods in terms of Continuous Ranked Probability Score (CRPS). CRPS is a commonly used evaluation metric in the field of probabilistic forecasting. The definition of CRPS is as follows:
$$
    \operatorname{CRPS}\left(F^f, F^0\right)=\int_{-\infty}^{+\infty}\left[F^f(x)-F^0(x)\right]^2 \mathrm{~d} x,
$$
where $F^f$ is predictive CDF and $F^0$ is real CDF. CRPS measures the differences between predictive distribution and real distribution. The lower the CRPS, the more accurate the probabilistic forecast results. In most cases, the real distribution is unknown, and we only know the observation value. If we have real observations of prediction steps $\xi_1, \xi_2, \ldots \ldots, \xi_n$, and the corresponding predictive distributions $F_1, F_2, \ldots \ldots, F_n$, CRPS can be estimated by the following equation:
$$
\begin{aligned}
\operatorname{CRPS}(F, \xi) & =\frac{1}{n} \sum_{i=1}^n \operatorname{crps}\left(F_i, \xi_i\right) \\
& =\frac{1}{n} \sum_{i=1}^n \int_{-\infty}^{+\infty}\left[F_i(x)-\varepsilon\left(x-\xi_i\right)\right]^2 \mathrm{~d} x,
\end{aligned}
$$
where $\varepsilon(t)$ is unit step function. There is no fixed threshold for how small the CRPS is to be a more accurate probabilistic forecast and it depends on different practical applications. The CRPS values under the same situation or data set are comparable.
\subsection{Result}

\begin{table}[!h]
\small
\centering
\caption{Average CRPS.}
  \label{tab: mean_crps}
    \begin{tabular}{llll}
      \toprule
      \diagbox{Method}{Dataset}& Tourism & Infant & Wiki      \\
        \midrule
        DeepAR-Hier                 & 23.854  & \textbf{8.468} & \textbf{3187.048}  \\
        Pure-DeepAR                 & 28.909  & 12.155 & 3841.136 \\
        Arima-Stack-MinT-Struct  & 25.036  & 11.532 & 18758.610  \\
        Arima-Stack-Bu-None      & 27.140   & 12.136 & 28156.813 \\
        Arima-Rank-MinT-Struct   & 23.364  & 15.140  & 17829.531 \\
        Arima-Rank-Bu-None       & 24.206  & 16.349 & 27377.863 \\
        Arima-Random-MinT-Struct & 25.329  & 11.527 & 18587.533 \\
        Arima-Random-Bu-None     & 27.138  & 12.161 & 27913.228 \\
        Ets-Stack-MinT-Struct    & 21.312  & 11.609 & 16312.947 \\
        Ets-Stack-Bu-None        & 26.589  & 12.996 & 23884.263 \\
        Ets-Rank-MinT-Struct     & \textbf{20.724}  & 15.563 & 15663.066 \\
        Ets-Rank-Bu-None         & 23.998  & 16.545 & 23162.240  \\
        Ets-Random-MinT-Struct   & 21.072  & 11.602 & 16386.476 \\
        Ets-Random-Bu-None       & 26.673  & 12.960  & 23055.768 \\
        \bottomrule
    \end{tabular}
\end{table}
Using the average CRPS of all series in one data set as the evaluation metric, the performance of each method is shown in Table \ref{tab: mean_crps}, where the minimum average CRPS value of each data set has been bolded. For Infant and Wiki data sets, DeepAR-Hier is the best approach compared with other probabilistic forecast reconciliation methods. For the Tourism data set, the performance of DeepAR-Hier is worse than Ets-Rank-MinT-Struct.

\begin{table}[!h]
\small
\centering
\caption{Average CRPS at each level.}
  \label{tab: mean_crps2}
    \begin{tabular}{cccr}
      \toprule
    Dataset     & Level & $\lambda$ & DeepAR-Hier \\
    \multirow{4}{1cm}{Tourism} & 1  & \multirow{4}{1cm}{0.006} & 1014.807   \\
        & 2  &       & 241.295     \\
        & 3  &       & 35.892      \\
        & 4  &       & \textbf{12.685}     \\
\multirow{4}{1cm}{Infant}  & 1  & \multirow{4}{1cm}{0.12}  & \textbf{32.280}     \\
        & 2  &       & \textbf{9.694}      \\
        & 3  &       & \textbf{6.366}          \\
\multirow{4}{1cm}{Wiki}    & 1  & \multirow{4}{1cm}{0.24}  & \textbf{31991.549}    \\
        & 2  &       & \textbf{6273.763}      \\
        & 3  &       & \textbf{2469.683}    \\
        & 4  &       & \textbf{2001.627}     \\

        \bottomrule
    \end{tabular}
\end{table}

We calculated the average CRPS at each level for the three data sets, i.e. the mean of CRPS values of all time series at a certain level. The results are shown in Table \ref{tab: mean_crps2}, where the $\lambda$ column gives the optimal $\lambda$ value determined on the validation set of each data set. The CRPS value in bold indicates that DeepAR-Hier is the best method at this level. It can be seen that for the Tourism data set, although DeepAR-Hier is not as good as other methods in terms of the average CRPS of all nodes, at the bottom level, the average CRPS of the DeepAR-Hier is the smallest. For the Wiki and Infant data set, DeepAR-Hier outperforms the other methods at all levels. In general, DeepAR-Hier is better than other probabilistic forecast reconciliation methods in most situations, but specifically, DeepAR-Hier can improve the probabilistic forecast accuracy of the bottom level, and performance at upper levels is not as good as that at lower levels.

\begin{figure}
  \begin{center}
    \begin{tabular}{c}
      \includegraphics[width=8cm]{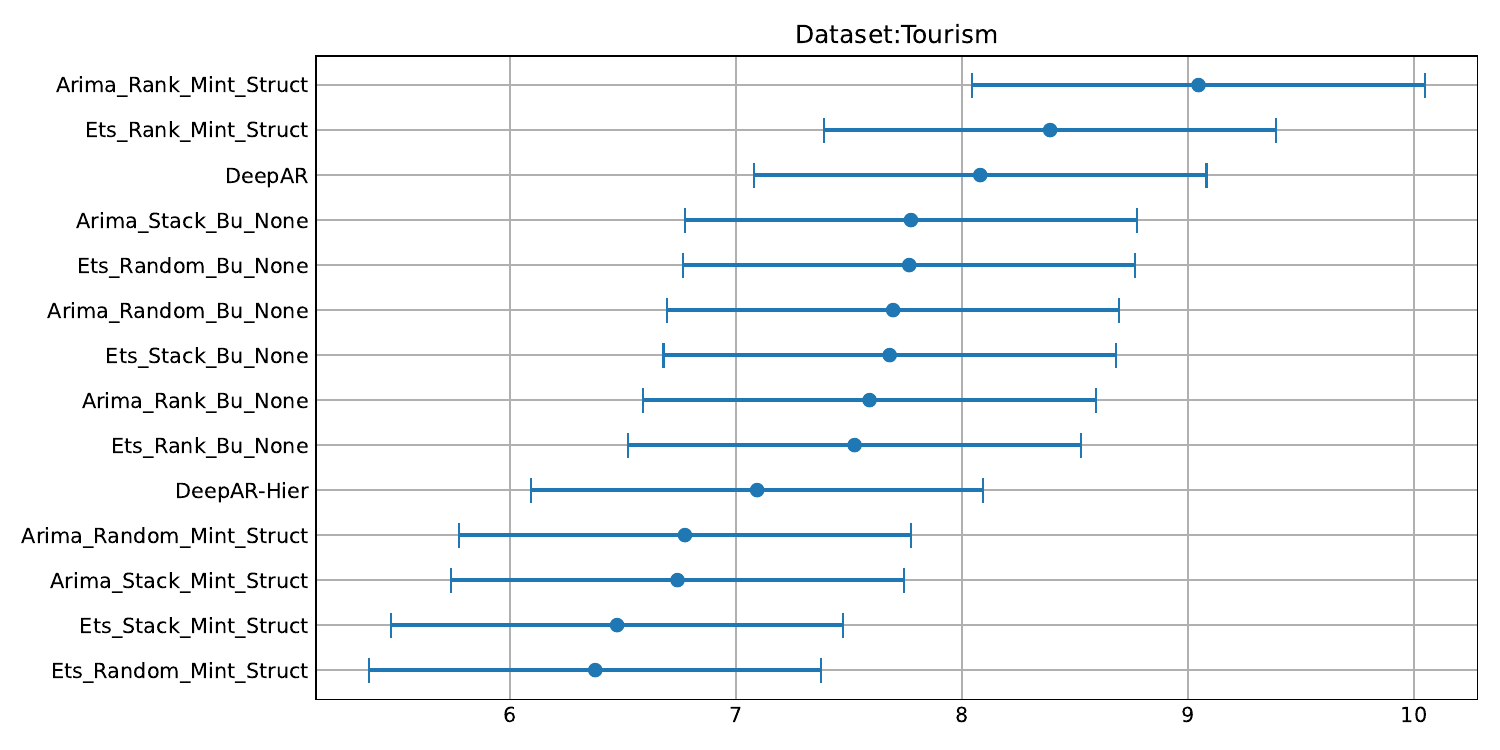}
    \end{tabular}
  \end{center}
  \caption
  { \label{MCB1}
    MCB result of Tourism dataset. }
\end{figure}

\begin{figure}
  \begin{center}
    \begin{tabular}{c}
      \includegraphics[width=8cm]{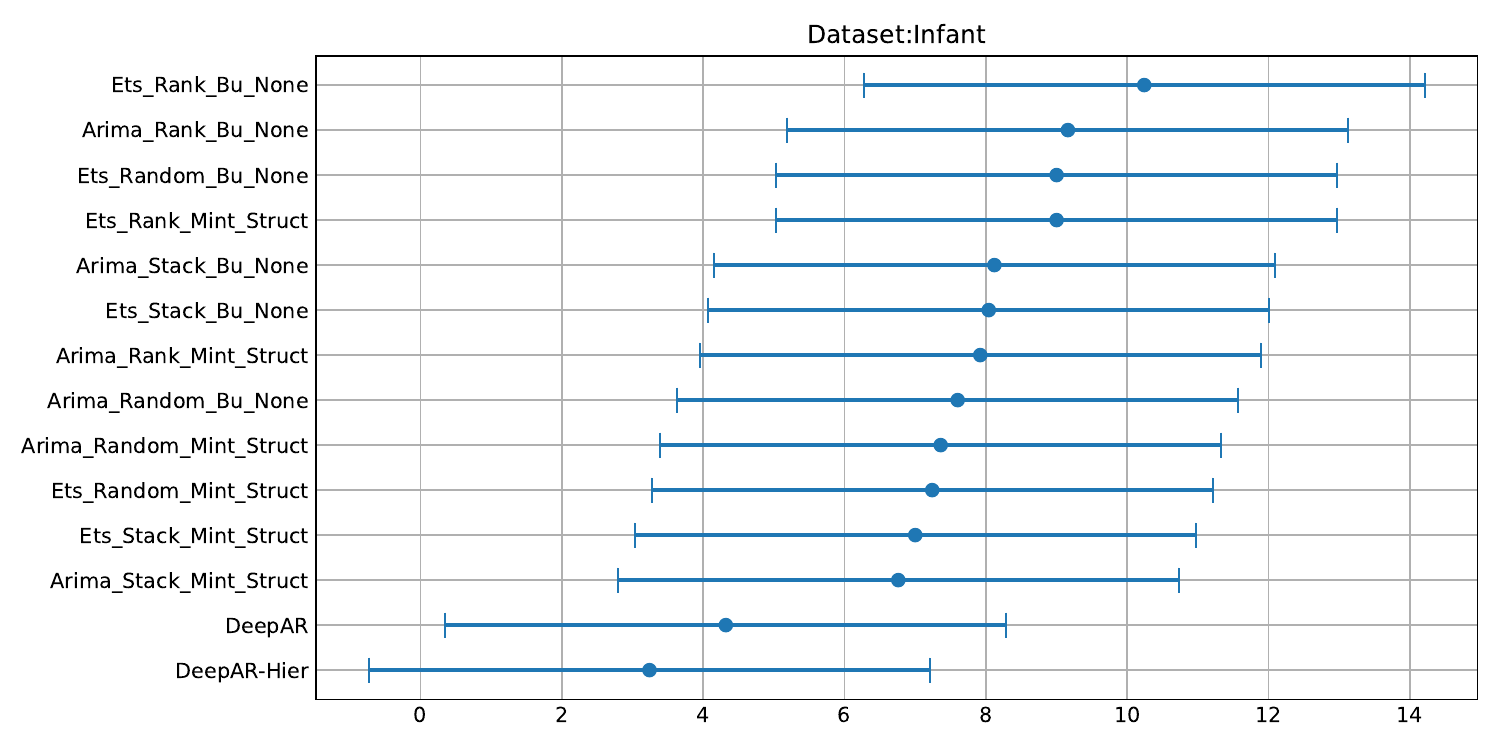}
    \end{tabular}
  \end{center}
  \caption
  { \label{MCB2}
    MCB result of Infant dataset. }
\end{figure}

\begin{figure}
  \begin{center}
    \begin{tabular}{c}
      \includegraphics[width=8cm]{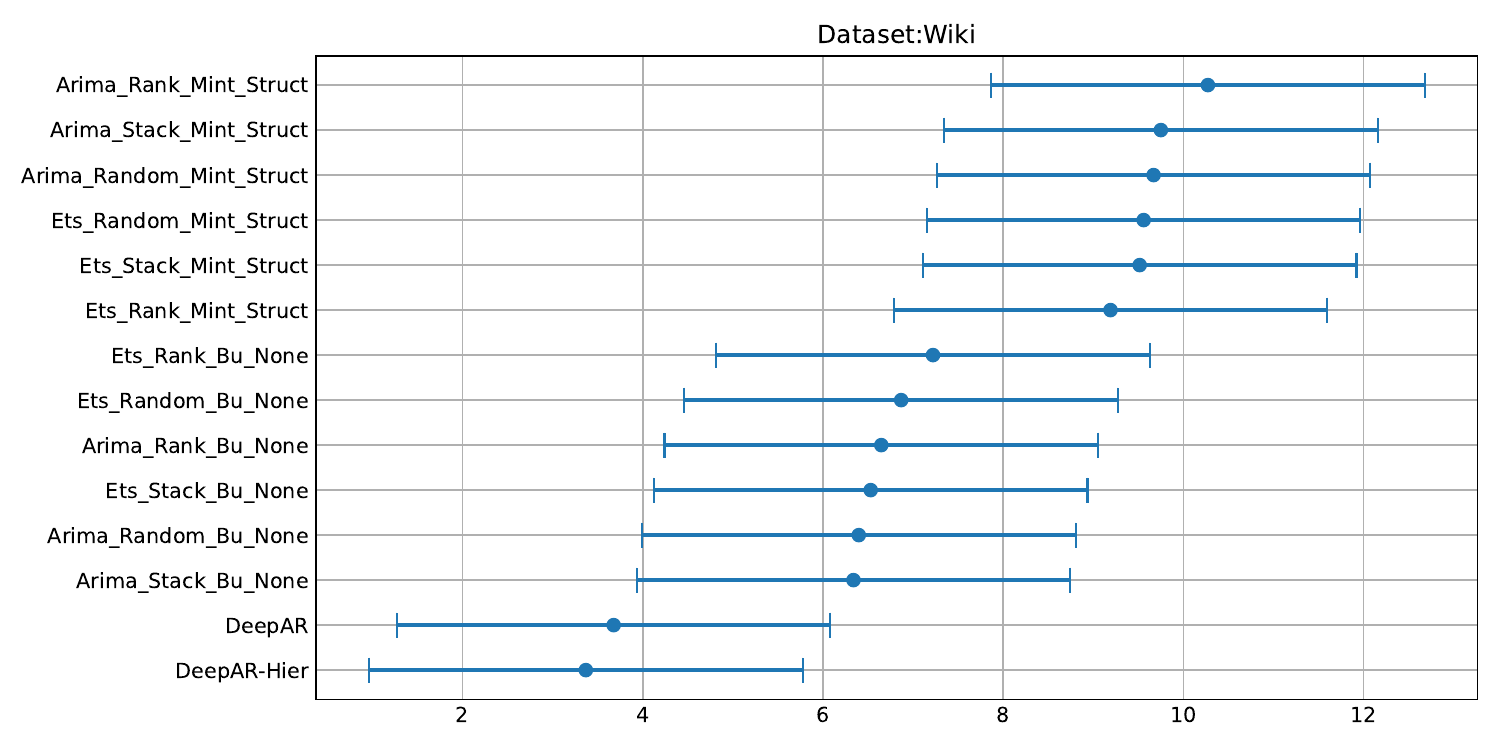}
    \end{tabular}
  \end{center}
  \caption
  { \label{MCB3}
    MCB result of Wiki dataset. }
\end{figure}

In order to evaluate the performance of each method more appropriately, we use multiple comparisons with the best (MCB) proposed by  Koning \emph{et al.} to test the statistical significance of the performance differences of multiple methods. MCB focuses on the average rank, i.e. the average rank of the evaluation metrics of all nodes \cite{KONING2005397}. MCB calculates the average rank and a confidence interval for each method to be compared. In Figure \ref{MCB1}, \ref{MCB2} and \ref{MCB3}, we present the average rank of each method and its confidence interval, and the results for each data set are shown in different panels. The best method for each data set is shown at the bottom of the panel, and the worst-performing method is shown at the top of each panel. If the intervals do not overlap, it indicates that the difference between the two methods in forecasting performance is statistically significant. It can be seen that for the Infant and Wiki datasets, the DeepAR-Hier method performs best. Although the average CRPS of DeepAR-Hier for the Tourism dataset is not the smallest, the forecast performance of DeepAR-Hier does not differ significantly from that of other methods because of interval overlapping.

\section{Conclusions and discussion}
\label{sec:conclusions}
In this paper, we propose a probabilistic forecast reconciliation method based on deep learning with Kullback-Leibler divergence regularization. We use a deep learning model to predict multivariate time series and introduce the Kullback-Leibler divergence regularization term into the loss function. This approach can fuse the prediction step and reconciliation step into a deep learning framework, instead of treating the reconciliation step as a fixed and hard post-processing step, making the reconciliation step more flexible and soft.

In our approach, the deep learning method is not specified, all global probabilistic forecasting models for multivariate time series based on deep learning can be used in our approach. We apply DeepAR to our proposed reconciliation method for probabilistic forecasting and conduct experiments on three hierarchical time series data sets. The experiment result shows that our approach is better than other probabilistic forecast reconciliation methods in most situations, but specifically, our approach can improve the probabilistic forecast accuracy of the bottom level, and performance at upper levels is not as good as that at lower levels.

There are several valuable directions worthy of further investigation. First, the approach we propose is aimed at cross-sectional data, in the future, we can try to apply the idea to temporal hierarchy and cross-temporal hierarchy. Second, DeepAR-Hier performs best on the Wikipedia page views data set, which is related to the fact that deep learning methods are suitable for large-volume data. In the future, deep learning frameworks suitable for small sample size data should be studied, and the minimum length of training time series should also be explored. It can be considered to conduct an experiment to test the trend of accuracy change relative to the training scale to determine the optimal training length. Finally, in our experiments, we assume that the predictive distribution is Gaussian distribution and we can try to replace the predictive distribution in the future.

\section*{Acknowledgments}
We thank the reviewers for helpful comments that improve the contents of this paper. This research was supported by the National Social Science Fund of China (22BTJ028).

\bibliographystyle{IEEEtran}
\bibliography{IEEEabrv,ref2}

\end{document}